\pdfoutput=1

\documentclass[11pt]{article}

\usepackage[review]{emnlp2021}

\usepackage{times}
\usepackage{latexsym}

\usepackage[T1]{fontenc}

\usepackage[utf8]{inputenc}

\usepackage{microtype}

\usepackage{scrextend} 
\usepackage{graphicx} 

\title{Construct a Sentence with Multiple Specified Words}

\author{Yuanliang Meng \\
  Nuance Communications Inc. / Burlington, MA \\
  \texttt{yuanliang.meng@nuance.com} \\
}

\begin{document}
\maketitle
\begin{abstract}
This paper demonstrates a task to finetune a BART model so it can construct a sentence from an arbitrary set of words, which used to be a difficult NLP task. The training task is making sentences with four words, but the trained model can generate sentences when fewer or more words are provided. The output sentences have high quality in general. The model can have some real-world applications, and this task can be used as an evaluation mechanism for any language model as well. 
\end{abstract}

\section{Introduction}

Making sentences from a given word, or a list of words, is a commonly used training and examining method in language pedagogy. 
For machines, nevertheless, people have long been wondering whether it is achievable. To the best of our knowledge, there does not exist any well-known tool for this purpose.
A powerful search engine may be able to locate sentences with one or two specified words, but the task becomes impossible with too many words: after all, such a sentence may not exist in any text available.

In recent years, Transformer-based self-supervised models have been springing up, achieving remarkable success in a variety of NLP tasks~\cite{vaswani2017attention,devlin-etal-2019-bert,liu2019roberta,yang2020xlnet,joshi-etal-2020-spanbert,brown2020language}. Therefore it is a good time to ask the question: can we now automatically generate sentences containing a given list of words?

BART~\cite{lewis2019bart} is a sequence-to-sequence model using reconstructing corrupted text as the training objective. It uses 5 specific denoising schemes (i.e. ``transformations"), 
and three of them are closely related to sentence-forming.

The token masking task of BART is literally the same as the one used in BERT~\cite{devlin-etal-2019-bert}. Random tokens are replaced with a mask, and the model needs to reconstruct them. The token deletion task randomly deletes tokens from text, without showing any place holder. The text infilling task can be viewed as a generalized token masking task. It replaces text spans with masks, or inserts masks in the text.
In all the schemes above, most tokens from the original text are still available as input, and they are more or less in their original positions. 
Our sentence-forming task, on the other hand, needs to creatively generate most words without any fixed context. The given set of words is in an arbitrary order. There is no ``correct answer" for the output, unlike in self-supervised learning where a reconstruction loss can be easily defined.

\section{Method}

Our problem can be formalized this way. Given a set of words $W=\{w_1, w_2,...,w_k\}$, make a 
well-formed sentence $S=[w'_1, w'_2,...,w'_n]$ using \emph{all of the words} i.e. $W\subset \textup{set}(S)$. $S$ here contains words, punctuation marks, and other legitimate tokens, if any. 


\subsection{Experiment setup}\label{subsection:experiment-setup}

We use the 324Mb europarl English monolingual data\footnote{https://www.statmt.org/wmt14/training-monolingual-europarl-v7/europarl-v7.en.gz}. It contains 2,218,201 lines of sentences. We only use sentences with 8$\sim$25 tokens (words and punctuation marks), and there are 1,171,662 of them. 70\% are used as training data, 10\% development data and 20\% test data. After the model configurations were determined, the development set joined the training set. In our sentence-forming task, the test data does not really provide the ``ground truth" as in many other seq2seq tasks, because the same set of words can make very different sentences. However, it is still important to make sure that training and evaluation do not overlap.


Within each sentence, $k$ words are randomly picked and shuffled as ``seeds". The seeds are the input to the model. We use a double underscore with spaces on both sides `` \_\_ " to separate the seeds, because it does not occur in a normal sentence, and thus will not complicate the task with its co-occurrence with words. 
When choosing the seeds, we also avoid words from a stopword list, because for the task to be more interesting, we want the seeds to be content words, instead of functional words such as determiners, prepositions, and auxiliaries. We also want to avoid some very high frequency words such as $very$ and $now$. 
Our stopword list is almost identical to the one used in NLTK, but NLTK stopwords include all the pronouns, while ours do not. The seeds also avoid numbers and non-word tokens.

As an example, the following is the first four sentences from our dev set (no cherry-picking). An example of their automatically sampled seeds is also shown in below.

\vspace{.4em}
{\footnotesize
\underline{Sentences:}
\begin{enumerate}
\item That article, however, must be implemented still further. \vspace{-0.5em}
\item I will deal with those two subjects separately: \vspace{-0.5em}
\item The clarification we seek serves the aim of securing legitimacy and transparency. \vspace{-0.5em}
\item That is exactly why the draft decision specifies that the political association should be active in at least one-third of the Member States. \vspace{-0.5em}
\end{enumerate}
\vspace{.4em}

\underline{Extracted seeds:}
\begin{enumerate}
\item implemented \_\_ article \_\_ still \_\_ must\vspace{-0.5em}
\item deal \_\_ two \_\_ separately \_\_ subjects\vspace{-0.5em}
\item seek \_\_ legitimacy \_\_ aim \_\_ securing\vspace{-0.5em}
\item Member \_\_ association \_\_ States \_\_ exactly\vspace{-0.5em}
\end{enumerate}
}
In our current experiment, we use $k=4$ to prepare training data. 
In test data, however, we use $k=2,4,6$. 
We also implemented two sampling methods for seed extraction. In one setting, each sentence in the training set is only used once, with four words being extracted and shuffled. In the other setting, we randomly sample and shuffle seeds from each sentence three times, and all the three input-output pairs are used as training data. 
Word tokens are case sensitive in all experiments. 

We use the Fairseq~\cite{ott-etal-2019-fairseq} framework for our experiments. The bart.large model\footnote{https://dl.fbaipublicfiles.com/fairseq/models/bart.large.tar.gz} is used as the pre-trained model. We use the label smoothed cross entropy loss~\cite{DBLP:journals/corr/PereyraTCKH17} with label smoothing value 0.1.
The models are trained on 8-GPU nodes, with 6144 maximal tokens in a batch per GPU. With three samples per sentence, the model is trained with 70 epochs. With one sample per sentence, 120 epochs are used.

\section{Results and Evaluation}

The results are evaluated on 4 test sets: europarl test data with 2, 4, 6 seeds, respectively, as well as external data.

The evaluation, naturally, contains two parts. On the one hand, we need to assure all seeds are used in the constructed sentences; and on the other hand, the sentences should be well-formed. The first criterion can be measured by the number of sentences missing one or more seeds. The second criterion, however, is challenging.
With some random check, we find there are very few if any syntactic errors in the output, and thus any automatic grammar checker would be unnecessary. They are also semantically acceptable for the most part. 
We eventually recruited human judges to evaluate a subset of output.


\subsection{Seed coverage evaluation}

\begin{table*}[h]
\centering
\resizebox{.7\textwidth}{!}{
\begin{tabular}{|l|c|c|c|c|}
\hline
\textbf{Measure} & \textbf{1-sample training,} &\textbf{3-sample training,} &\textbf{3-sample training,} &\textbf{3-sample training,}\\
 & \textbf{4 seeds test (\%)} &\textbf{4 seeds test (\%)} &\textbf{2 seeds test (\%)} &\textbf{6 seeds test (\%)}\\
\hline
Miss 6 &- &- &- &0.0 \\
Miss 5 &- &- &- &0.0 \\
Miss 4 &0.0 &0.0 &- &0.1 \\
Miss 3 &0.1 &0.0 &- &1.1 \\
Miss 2 &1.2 &0.4 &0.1 &11.4 \\
Miss 1 &13.8 &5.0 &2.4 &39.5 \\
Perfect &84.9 &94.6 &97.5 &48.0 \\
\hline
\end{tabular}
}
\caption{Coverage evaluation results. 3-sample training mean each sentence in the training set is used three times to sample seeds. Training data always has 4 seeds for each sentence, but test data may have 2, 4 or 6.}
\label{tab:results}
\end{table*}

As Table~\ref{tab:results} shows, sampling seeds three times to create the training set has a clear advantage over sampling only once. Given the training data, the difference cannot be explained by the number of model updates alone, because we did try longer epochs with the 1-sample case. 
It would be interesting to investigate if the same is true with a much larger dataset.

As a proof of concept, all models are trained with 4 seeds only. However, as the results suggest, when we expand the test data to 6 seeds the model is still able to use all seeds in nearly half of the cases. In fact, it successfully constructs sentences with 5$\sim$6 seeds in $87.5\%$ of the cases. Clearly it has the capability to generalize the task with respect to the number of seeds. It should be noted that with more given words, it becomes more difficult to make a sentence, which is true for humans as well.

Another interesting phenomenon is, when there are fewer seeds, the model tends to generate shorter, dull sentences too. This is easy to explain: shorter sequences with high frequency tokens can get higher beam scores. 

\begin{table}[h]
\centering
\resizebox{0.7\columnwidth}{!}{
\begin{tabular}{|l|c|c|c||c|}
\hline
 \textbf{\# seeds}& \textbf{2} &\textbf{4} &\textbf{6} &\textbf{test set}\\
\hline
avg \# words &8.2 &12.9 &15.0 &16.6\\
avg \# chars &42.3 &74.1 &86.7 &99.0 \\
\hline
\end{tabular}
}
\caption{Average number of words and characters per generated sentence, with different number of seeds in test data. We use the 3-sample model to generate the results here.}
\label{tab:lengths}
\end{table}

Table~\ref{tab:lengths} shows the average lengths of generated sentences, with different number of seeds in the experiment. Human written sentences in our test set are longer in general. Two factors are relevant here: (1) Making sentences is different from writing articles, and it is reasonable to avoid anything extra;
(2) This corpus uses news articles, and the formality leans to lengthy expressions.

In regard to the extreme cases where the system encounters big failures, there are 13 instances where the model uses none of the 4 seeds. One example is with the seeds \textit{agus, Eireann, Uachtaranacht, Comhphobail}, not a single English word. Another case has all Spanish names, and so on. When fewer words are missing, it often involves abbreviations. For example, when ``SPD" is used as a seed, presumably referring to the Social Democratic Party in that sentence, the generated sentence contains ``SPDC".
As noted before, our criterion is case sensitive. If a lower-case seed word exists in the output sentence, but it becomes capitalized, it still counts as missing.

\subsection{Quality evaluation}

The generated sentences are usually desirable. The following is the results from the seeds shown in Section~\ref{subsection:experiment-setup} (again, no cherry-picking):

\vspace{.4em}
{\footnotesize
\underline{System output:}
\begin{enumerate}
\item However, this article must still be implemented. \vspace{-.5em}
\item As they deal with the same subject, the two subjects will be dealt with separately. \vspace{-.5em}
\item  The aim of this report is to seek greater legitimacy in securing the necessary funding. \vspace{-.5em}
\item This is exactly what the association agreements with the Member States are about. \vspace{-.5em}
\end{enumerate}
}



In order to evaluate the naturalness of the sentences, we recruited 6 native speakers of English with college education. We listed our output sentences in pairs with the original sentences used to extract seeds. Then we randomly selected 100 pairs in which the lengths of the two sentences do not differ by more than 3 tokens.
For each pair, the 6 participants needed to vote which one is written by a \textit{human}. They were also allowed to answer ``I cannot tell". Eventually 1 of the 100 pairs happens to contain identical sentences. Among the remaining 99 pairs, 50 human written sentences get more votes, while 40 machine generated sentences get more votes. Results are shown in Table~\ref{tab:votes}. With a close look, \emph{none} of the 99 human written sentences passed unanimously i.e. at least 1 participant voted for the machine generated one, in any pair.

\subsection{Seeds across sentences}
So far the seeds in the test set are extracted from true sentences. Moreover, the training and test sets are more or less in similar genres, following similar styles. It is necessary to examine without further training, whether the model works on more random collections of words, or words from a very different genre.
We constructed 2,000 4-seed sets over the whole text of \textit{Pride and Prejudice}\footnote{https://www.gutenberg.org/files/1342/1342-0.txt}, which has about 125k words, and then generated sentences from these seeds. Similarly, we pick the first 1/3 chunk of our test set, which has about 130k words, and sample 4-seeds sets over the whole chunk of data. Now any 4 seeds in a group can be from different sentences, so the task may become more difficult. 

\begin{table}[h]
\centering
\resizebox{0.6\columnwidth}{!}{
\begin{tabular}{|l|c|c|}
\hline
\textbf{Measure} & \textbf{europarl (\%)} &\textbf{P \& P (\%)} \\

\hline
Miss 4 &0.0 &0.0  \\
Miss 3 &0.0 &0.5\\
Miss 2 &0.6 &5.1 \\
Miss 1 &8.4  &25.0 \\
Perfect &91.0 & 69.5\\
\hline
\end{tabular}
}
\caption{Results from seeds across sentences, and from different documents. Here europarl means the first 1/3 chunk of our europarl test data,and P\&P means \textit{Pride and Prejudice}}
\label{tab:documents}
\end{table}
As illustrated in Table~\ref{tab:documents}, the performance does not drop much with the europarl dataset, but it is much lower on P\&P data. There is no doubt that news articles from recent years are very different from a novel written in 1813. With a close look at the seeds extracted from P\&P data, we see lots of person names. With a calculation, P\&P has 2.2 uppercase characters per seed set (4 seeds), while europarl only has 0.80. From another perspective, while pre-training models like BART are successful in many downstream NLP tasks, its capability to handle proper names may still be unsatisfactory.

We also recruited human judges to evaluate the naturalness. Since the seeds are sampled over a document, not within sentences, there is no ``paired" sentence to compared to. However, we randomly selected real sentences from test data to pair with the generated sentences, and the participants were still asked to identify the human written sentences. Two human judges participated in this task. As Table~\ref{tab:votes} shows, for the task involving europarl seeds, 41 out of 100 human written sentences received both votes, while 10 machine generated sentences received both votes, and 49 cases are ambiguous. For the task involving P\&P seeds, 47 out of 100 human written sentences received both votes, while 12 machine generated sentences received both votes, and 41 are ambiguous.

Therefore, although the machine generated sentences are less likely to be a clear winner, they are clear ``losers" in less than half of the cases only. 
It should be mentioned that the sentences which do not get votes are not necessarily unacceptable. Our task forces human judges to pick one out of two, which is a very strict criterion. 

\begin{table}[h]
\centering
\resizebox{1.0\columnwidth}{!}{
\begin{tabular}{|l|c|c|c|}
\hline
\textbf{seeds for} & \textbf{europarl} & \textbf{europarl} &\textbf{P \& P} \\
\textbf{sys output} & \textbf{within-sentence} & \textbf{cross-sentence} & \\

\hline
human wins &50 &41 &47  \\
machine wins &40 &10 &12\\
tied &9 &49 &41\\
\hline
\# judges &6  &2 &2\\
\hline
\end{tabular}
}
\caption{Human evaluation results of three paired identification tasks.}
\label{tab:votes}
\end{table}\vspace{-.8em}

\section{Colorless green ideas sleep furiously}
In theoretical linguistics, some argue that speakers can make sense of syntactically well-formed but semantically anomalous sentences. \newcite{Chomsky57a} used a famous example, ``Colorless green ideas sleep furiously'', to illustrate it. He argues it demonstrates the distinction between syntax and semantics, and thus (certain) probabilistic models of grammar are inadequate.
However, \newcite{Pereira2000} trained a Markov model with newspaper text and showed that under this model, this sentence has a much higher probability than its ``ungrammatical'' counterparts consisting of the same words. 

We are also interested in seeing if our model can generate similar sentences. 
Our initial attempt is to use exactly the same words in Chomsky's example as seeds to generate sentences. However, it turns out \emph{colorless} and \emph{furiously} are both very low-frequency words in our training data (and probably in pretraining data of BART too), so making sentence with them is not successful. Therefore we made some changes and considered the sentence ``tasteless sweet ideas sleep emotionally''. These five lowercase words are used as seeds, with randomized orders, and the model trained on europarl data (with 4 seeds) is applied to generate sentences. Two examples of the results are shown in below.

{\footnotesize
\begin{enumerate}
\item They are not so sweet as tasteless ideas that they cannot sleep emotionally. \vspace{-0.5em}
\item These ideas often sleep very emotionally, because they are very sweet.\vspace{-0.5em}
\end{enumerate}
}
As we can see, the generated sentences look grammatical, but not meaningful. There does not need to be any innate, structured syntactic rules in the model to perform this way. Statistical models based on co-occurrences in natural language can easily learn to produce such patterns.

\section{Conclusion and discussion}
We have designed a method to finetune a BART model to construct sentences using an arbitrary set of words. The results have desirable quality, as demonstrated by seed coverage statistics and human evaluation. 
The model can also generate syntactically well-formed, but semantically anomalous sentences, although it has never been trained on such data.

Self-supervised masked language models have shown very impressive performances on a variety of NLP tasks, and more pre-training mechanisms are still being introduced. Sentence-forming is a commonly used technique for training and testing human language learners. Even if it is difficult to be directly implemented in self-supervised learning, it may be considered for model evaluation. 

\section*{Acknowledgments}
This is an independent research project.

\bibliography{anthology,custom}
\bibliographystyle{acl_natbib}

\appendix



\end{document}